**Paper Code: DSIP-024**                                                                                           Oral

# A NOVEL SCHEME FOR BINARIZATION OF VEHICLE IMAGES USING HIERARCHICAL HISTOGRAM EQUALIZATION TECHNIQUE

Satadal Saha [1], Subhadip Basu [2]*, Mita Nasipuri [2], Dipak Kumar Basu [2]

[1] CSE Department, MCKV Institute of Engineering, Howrah, India
[2] CSE Department, Jadavpur University, Kolkata, India
* Corresponding author. Email: subhadip@ieee.org

**Abstract:** *Automatic License Plate Recognition system is a challenging area of research now-a-days and binarization is an integral and most important part of it. In case of a real life scenario, most of existing methods fail to properly binarize the image of a vehicle in a congested road, captured through a CCD camera. In the current work we have applied histogram equalization technique over the complete image and also over different hierarchy of image partitioning. A novel scheme is formulated for giving the membership value to each pixel for each hierarchy of histogram equalization. Then the image is binarized depending on the net membership value of each pixel. The technique is exhaustively evaluated on the vehicle image dataset as well as the license plate dataset, giving satisfactory performances.*

## 1. INTRODUCTION

Automatic license plate localisation from vehicle images and subsequent recognition of alphanumeric characters from the localized license plates has long been an active area of research because of its potential use in various applications in our daily life. The main objective of such systems is to localize the license plate region(s) from the vehicle images captured through a road-side camera and interpret them using an Optical Character Recognition (OCR) system. The localization of license plate is itself a challenging task due to its wide variations in size, shape, colour, texture, spatial orientation and noise.

Usually such system consists of three main parts: license plate localisation, character segmentation and character recognition. Out of these license plate localisation is the most important and also most difficult task compared to the other two. The main way through which a license plate can be localised depends on its distinguished colour, contrast and the character pattern of it. To extract proper information of the license plate, the binarization of the image is an utmost requirement. In a controlled environment and with accurate acquisition of image, the binarization remains more or less easier. But in a practical outdoor scenario the image is heavily corrupted by variation of different uncontrolled impairments, such as lighting conditions, colour/ grey scale content, wind turbulence, vibration of camera etc. Finding a generalised way of binarization becomes a trivial task in these conditions.

Binarization is the method of converting a gray scale image (popularly known as multi-tone image) into a black-and-white image (popularly known as two-tone image). This conversion is based on finding a threshold gray value and deciding whether a pixel having a particular gray value is to be converted to black or white. Usually within an image the pixels having gray value greater than the threshold is transformed to white and the pixels having gray value lesser than the threshold is transformed to black. Binarization has been the area of research for last fifteen years or so, mainly to find a single threshold value or a set of threshold values for converting a gray scale image into a binary image. Most of the algorithms till developed are of generic type using statistical parameters computed over the image with or without using local information or special content within the image.

The most convenient and primitive method is to find a global threshold for the whole image and binarize the image using the single threshold. In this technique, the local variations are actually suppressed or lost, though they may have important contribution towards the information content within it. On the other hand, in case of determining the threshold locally, a window is used around a pixel and threshold value is calculated for the window. Now depending on whether the threshold is to be used for the center pixel of the window or for all the pixels in the window, the binarization is done on pixel-by-pixel basis, where each pixel may have a calculated threshold value, or on region-by-region basis where all pixels in a region or window have same threshold value.

The major contribution of research for binarization is to recover or extract information from a degraded document image. Otsu [1] developed a method based on gray level histogram and maximizes the intra-class variance to total variance. Sauvola [2] developed an algorithm for text and picture segmentation within an image and binarized the image using local threshold. Gatos [3] used Wiener filter and Sauvola's adaptive binarization method. In the work presented in [4] also Sauvola's adaptive

*International Conference on Computer, Communication, Control and Information Technology (C³IT 2009)*

thresholding is used for binarization. Valverde [5] binariesd the image using Niblack's technique. A slight modification of Niblack's method is done in [6] by Zhang.

The main objective of the present work is to localize the license plate which is nothing but the text region within the image. We have applied histogram equalization method over the whole image globally and also over different hierarchy of image partitions. A novel scheme is formulated for giving the membership value to each pixel for each level of histogram equalization. Then the image is binarized depending on the net membership value of each pixel.

## 2. COLLECTION OF DATASET

The dataset for the current work has been constructed as a part of a demonstration project on Automated Red Light Violation Detection system for a Government traffic monitoring authority of a major metro city in India. An important road crossing in Kolkata is selected for collection of dataset. Three surveillance cameras were installed at the crossing at a height of around ten meters from the road surface. The major objective was to capture the images of red signal violating vehicles only. To serve the purpose, all the surveillance cameras were synchronized with the traffic signaling system such that the cameras capture the snapshots only when the traffic signal for a particular lane is turned RED. All the cameras were focused on the Stop-Line to capture frontal images of vehicles violating the Stop-Line on a RED traffic signal.

The complete image dataset comprises of more than 15,000 surveillance video snapshots, captured over several days/nights in an unconstrained environment accommodating varying outdoor lighting conditions, pollution levels, wind turbulences and vibrations of the camera. 24-bit color bitmaps were captured through CCD cameras with a frame rate of 25 fps and resolution of 704 x 576 pixels. It is seen that in many images there is no license plate, or partial license plate. For the current experiment, we have identified 2428 images with complete license plate regions appearing in different orientations in the image frame.

## 3. THE PRESENT WORK

It is already described in the previous section that true color video snapshots of resolution 704 × 576 pixels were captured through multiple surveillance cameras over day and night with embedded noise and huge variations in image quality.

### 3.1. Gray scale conversion

From the 24-bit color value of each pixel the R, G and B components are separated and the 8-bit gray value is calculated using the formula:

*gray(i, j) = 0.59 \* R(i, j) + 0. 30 \* G(i, j) + 0. 11 \* B(i, j)*   (1)

where, *(i,j)* indicates the location of a pixel.

### 3.2. Median filtering

Median filter is a non-linear filter, which replaces the gray value of a pixel by the median of the gray values of its neighbors. In this work, median filter having 3 × 3 mask is applied over the gray scale image to get eight neighbors of a pixel and their corresponding gray values. This operation removes salt-and-peeper noise from the image.

### 3.3. Histogram equalization

Contrast of each image is enhanced through histogram equalization technique, as discussed in [7]. Total 256 numbers of gray levels (from 0 to 255) are used for stretching the contrast. Let total number of pixels in the image is N and the number of pixels having gray level k be nk. Then the probability of occurrence of gray level k is, Pk = nk / N. The stretched gray level Sk is calculated using the cumulative frequency of occurrence of the gray level k in the original image using the formula:

$$S_k = \sum_{j=0}^{k} \frac{n_j}{N} \times 255 \qquad (2)$$

where, 255 indicates the maximum gray level in the enhanced image. This $S_k$ divided by 255 results an enhanced fractional gray value *($f_g$=$S_k$/255)* of a pixel in the range 0 to 1 and gives the likeliness of a pixel to be white. A fraction close to 1 indicates that the method suggests the pixel to be white. On the other hand a fraction close to 0 indicates that the method suggests the pixel to be black. Application of histogram equalization method over the whole image gives $0^{th}$ level membership value *($f_{mem}(i,j,0)$= $f_g$)* of pixel *(i,j)*.

### 3.4. Hierarchical partitioning of image and application of histogram equalization in each image partitions

Four equal segments of the image are generated by partitioning the image midway both width wise and height wise and histogram equalization method is applied in each segment separately over the original data. This provides 1st level membership value (fmem(i,j,1)) of pixel (i,j). Likewise each of the 1st level segments is divided into four equal segments and histogram equalization method is applied over each sub-segment to get the 2nd level membership value (fmem(i,j,2)) of pixel (i,j). This process goes on upto nth sub-segment and nth level membership value (fmem(i,j,n)) is obtained for pixel (i,j).

### 3.5. Combining membership values of each pixel

As discussed in the last subsection, for each level of contrast enhancement through histogram equalization, each pixel gets an enhanced fractional gray value leading to a membership value (fmem(i,j,level)) indicating that whether the said pixel is closer to white or black. These membership values for each pixel are combined to get a net membership value for each pixel. Now as the global scenario has a little effect over the local variations

*International Conference on Computer, Communication, Control and Information Technology (C³IT 2009)*

during the process of combination the local membership values are given more weightage over global membership values. Keeping this in mind the net membership value for pixel (i,j) is calculated using the formula:

$$f_{net}(i,j,level) = \frac{\sum_{level=0}^{n} f_{mem}(i,j,level)(level+1)^2}{\sum_{level=0}^{n}(level+1)^2} \quad (3)$$

where, the summation, $\Sigma$, is over the levels 0 to n. Thus during the process each pixel gets a net membership value in which it retains the global as well as the local information of the variation of gray level.

### 3.6. Binarization

As discussed in the last two subsections, for each pixel a net membership value is obtained, which is compared with the threshold value for binarization. As the image is histogram equalized image, the mean gray level lies in the middle position of the gray scale range (0 to 255). Thus 0.5 is chosen as threshold and any pixel having net membership value greater than 0.5 is converted to white, otherwise it is converted to black.

### 4. EXPERIMENTAL RESULTS

As discussed in Section 2, about 15000 24-bit color bitmaps were captured through CCD cameras with a frame rate of 25 fps and resolution of 704 x 576 pixels. They are converted to gray scale and the algorithm is applied on 2428 number of gray scale images. Our algorithm finds the membership value of each pixel for each level of localization. For the purpose of full image binarization, we have used level 2 to 8 and for binarizing license plate only we have used level 0 to 3. For the same set of images we have used Otsu's method of binarization as a benchmark and compared the two sets of result.

Fig. 1(a-h) shows a sample set of grey-scale vehicle images over which the proposed binarization algorithm and Otsu's algorithm are applied upon.

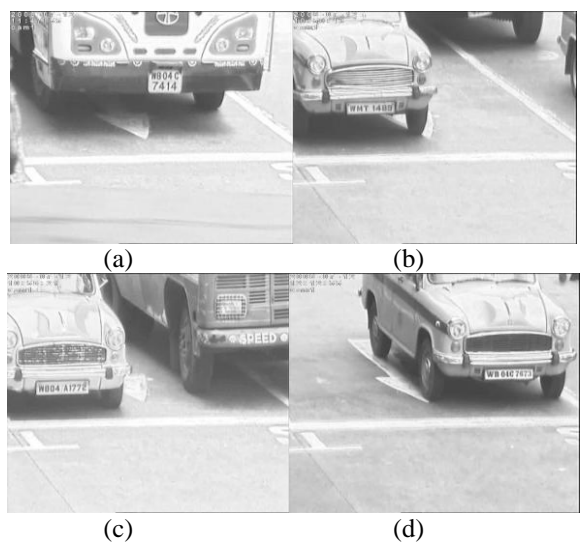

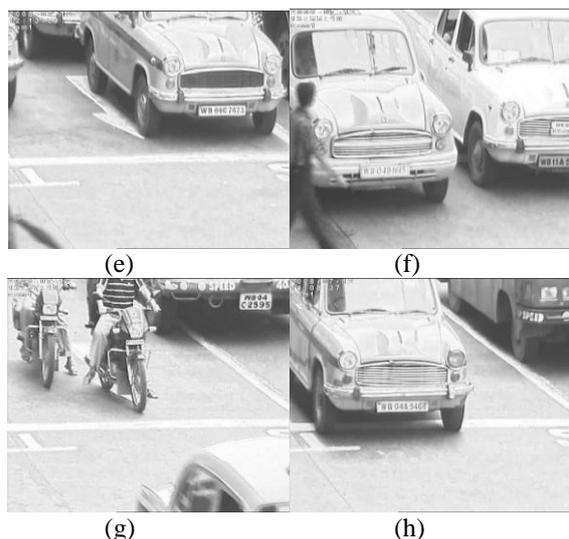

**Fig. 1(a-h).** Sample grey-scale vehicle images, considered for the current experiment

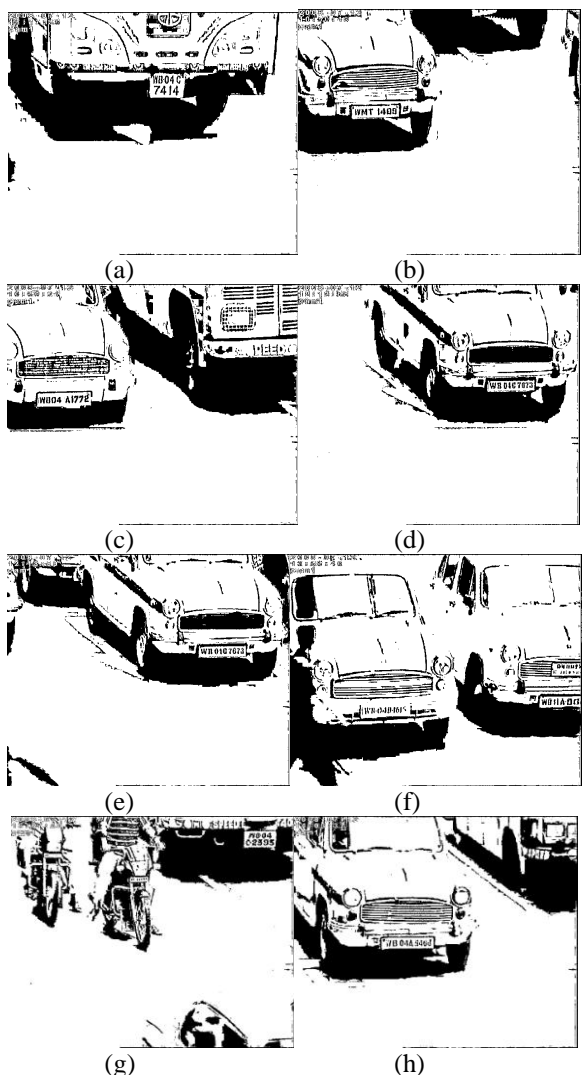

**Fig. 2(a-h).** Binarization result of our designed algorithm on the sample set of images shown in Fig. 1.

273

*International Conference on Computer, Communication, Control and Information Technology (C³IT 2009)*

Figures 2(a-h) and 3(a-h) show the binarized vehicle images as a result of application of our algorithm and Otsu's algorithm respectively on the sample set of images, as shown in fig. 1(a-h).

Comparing Fig. 2 and Fig. 3, it can be observed that our proposed algorithm results in a clearer, proper and expected binarized images rather than Otsu's method.

Fig. 4 shows some sample license plate regions manually cut out from the vehicle images. Both our proposed algorithm and Otsu's algorithm are applied upon these images.

Figures 5 and 6 show the binarized license plate images as a result of application of our algorithm and Otsu's algorithm respectively when applied on the images shown in Fig. 4.

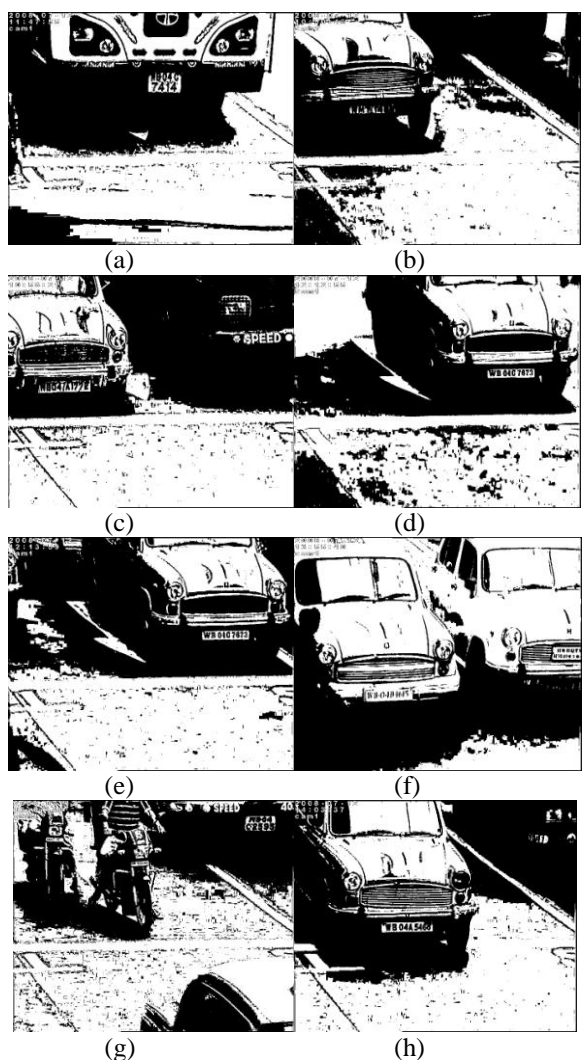

**Fig. 3(a-h). Binarization result of Otsu's algorithm on the sample set of images shown in Fig. 1.**

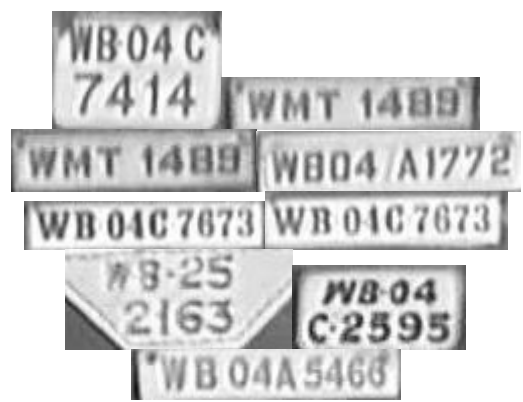

**Fig. 4. Sample Grey-scale images of vehicle license plates.**

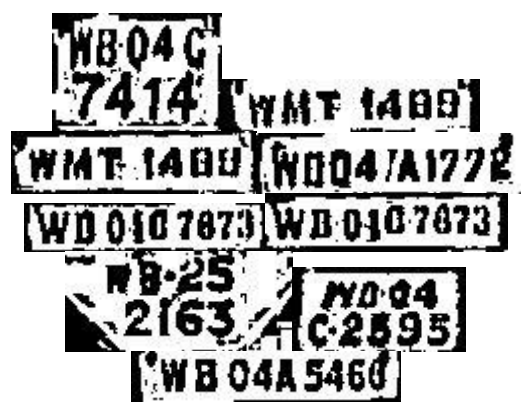

**Fig. 5. Binarization result of our designed algorithm on the sample set of vehicle license plate images shown in** Fig. 4.

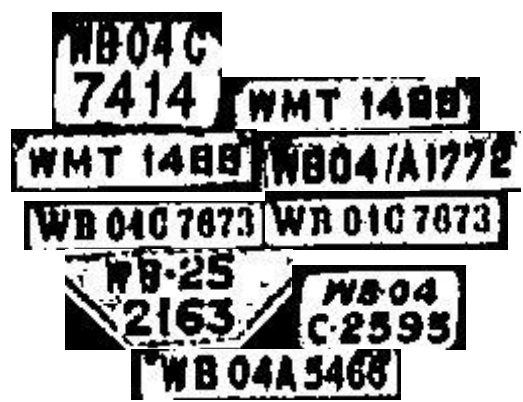

**Fig. 6. Binarization result of Otsu's algorithm on the sample set of vehicle license plate images shown in Fig. 4.**

Comparing Fig. 5 and Fig. 6 it is seen that the finer details are retained with our proposed method resulting in a thinner and noise like output compared to Otsu's method.

## 5. CONCLUSION

In the present work, we have developed an efficient way of binarizing a vehicle image captured in a practical outdoor scenario. We have applied Otsu's binarization algorithm over the images but our method seems to be more effective in binarizing the image. The algorithm is applied on a dataset of 2428 natural traffic images and satisfactory result is obtained in most cases. After localization of license plate the characters are to be segmented and recognize. During this segmentation phase also binarization is done again over the original data of localized license plate only. Our algorithm performs satisfactorily well on this reduced data space also.

## ACKNOWLEDGEMENT

Authors are thankful to the CMATER and the SRUVM project, C.S.E. Department, Jadavpur University, for providing necessary infrastructural facilities during the progress of the work. One of the authors, Mr. S. Saha, is thankful to the authorities of MCKV Institute of Engineering for kindly permitting him to carry on the research work.